\pdfoutput=1

\documentclass[11pt]{article}
\usepackage{flushend}
\usepackage[]{ACL2023}
\usepackage{pbox}
\usepackage{amsmath}
\usepackage{caption}
\usepackage{float}
\usepackage[labelformat=simple]{subcaption}

\usepackage{color,soul}
\usepackage{colortbl}
\usepackage{color, xcolor}
\usepackage{multirow}

\newcommand{\basebase}{T5\textsubscript{\tiny BASE}\xspace}
\newcommand{\basexlarge}{T5\textsubscript{\tiny 3B}\xspace}
\newcommand{\robertabase}{RoBERTa\textsubscript{\tiny BASE}\xspace}

\newcommand{\propet}{\textsc{ProPetl}\xspace}
\newcommand{\propetadapter}{\textsc{ProPetl}\textsubscript{\tiny Adapter}\xspace}
\newcommand{\propetprefix}{\textsc{ProPetl}\textsubscript{\tiny Prefix}\xspace}
\newcommand{\propetlora}{\textsc{ProPetl}\textsubscript{\tiny LoRA}\xspace}
\DeclareMathOperator*{\softmax}{softmax}
\DeclareMathOperator*{\attention}{Attention}
\usepackage{graphicx} 

\usepackage{times}
\usepackage{latexsym}
\usepackage{amsfonts} 
\usepackage{multirow}
\usepackage[english]{babel}

\usepackage{booktabs}
\usepackage{stfloats}
\usepackage{caption}
\usepackage{float} 

\usepackage[linesnumbered,ruled,vlined]{algorithm2e}
\SetKwComment{Comment}{/* }{ */}
\usepackage[T1]{fontenc}
\usepackage[utf8]{inputenc}

\usepackage{microtype}
\usepackage{amsmath}

\usepackage{inconsolata}

\title{One Network, Many Masks: \\Towards More Parameter-Efficient Transfer Learning}

\author{
 Guangtao Zeng$^{*}$ \quad  Peiyuan Zhang$^{*}$ \quad Wei Lu
 \\
StatNLP Research Group\\
Singapore University of Technology and Design\\
 \texttt{guangtao\_zeng@mymail.sutd.edu.sg, \{peiyuan\_zhang, luwei\}@sutd.edu.sg}
}

\begin{document}
\maketitle
\renewcommand{\thefootnote}{\fnsymbol{footnote}}
\footnotetext[1]{The first two authors contributed equally.}
\renewcommand{\thefootnote}{\arabic{footnote}}
\begin{abstract}
Fine-tuning pre-trained language models for multiple tasks tends to be expensive in terms of storage.
\textcolor{black}{To mitigate this}, parameter-efficient transfer learning (PETL) methods have been proposed to address this issue, but they still require a significant number of parameters and storage when being applied to broader ranges of tasks.
To achieve even greater storage reduction, we propose \propet, a novel method that enables efficient sharing of a single \textcolor{black}{PETL module which we call prototype network (e.g., adapter, LoRA, and prefix-tuning)} across layers and tasks.
We then learn binary masks to select different sub-networks from the shared prototype network and apply them as PETL modules into different layers. We find that the binary masks can determine crucial information from the network, which is often ignored in previous studies.
Our work can also be seen as a type of pruning method, where we find that overparameterization also exists in the seemingly small PETL modules.
We evaluate \propet on various downstream tasks and show that it can outperform other PETL methods  with approximately $10\%$ of the parameter storage required by the latter.\footnote{Our code is available at \url{https://github.com/ChaosCodes/ProPETL}.}

\end{abstract}

\section{Introduction}
\begin{figure}[t!]
    \centering
    \includegraphics[width=0.44\textwidth]{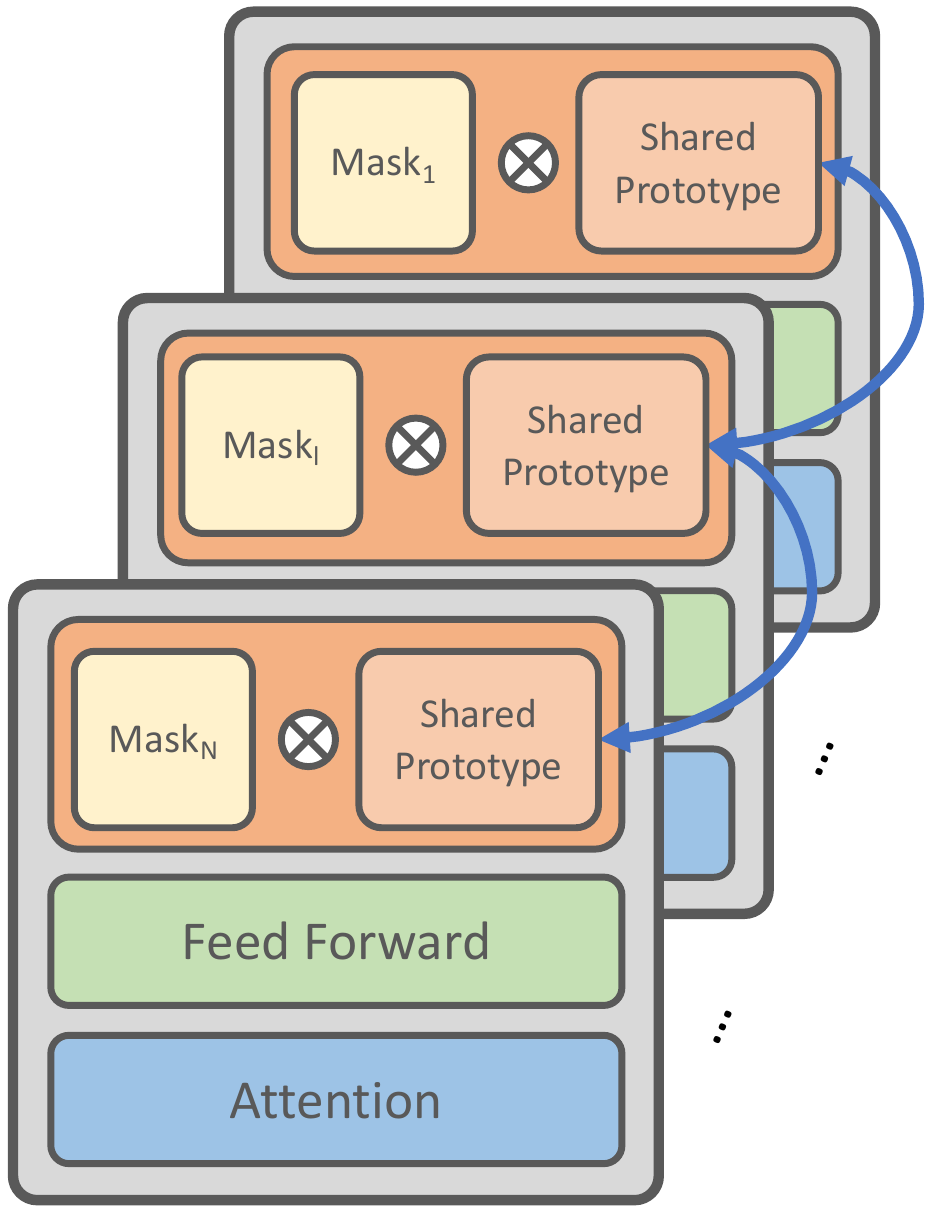}
    \caption{An illustration of our \propetadapter model. Note that \propet is orthogonal to the specific PETL architectures. LoRA and prefix-tuning are also implemented in our framework.}
    \label{fig:shareadapter}

\end{figure}
With the release and wide application of numerous pre-trained language models (PLMs)~\cite{DevlinCLT19, roberta}, pre-training and subsequently fine-tuning them becomes prevalent in natural language processing (NLP). This yields good performance in many downstream tasks.
However, such a paradigm requires the entire model to be updated and saved after fine-tuning. As PLMs grow in size, traditional fine-tuning becomes costly in storage, limiting the application to multi-task scenarios.
In order to ameliorate this issue, many Parameter-Efficient Transfer Learning (PETL) methods have been proposed ~\cite{HoulsbyGJMLGAG19, LiL20, HuSWALWWC22}.
Rather than fine-tuning the entire model, they introduce new parameters and only fine-tune those additional parameters on downstream tasks, which drastically reduces the  \textcolor{black}{storage of the parameters required by}  each task.
However, they still require a significant number of parameters when more tasks are considered. 

In this paper, we continue this line of research and target \textcolor{black}{using even less storage per task}.  
We observe that recent advancements in the field of PETL focus on finding better ways to apply the additional parameters, 
 such as the adapter module after each feed-forward layer \cite{HoulsbyGJMLGAG19, PfeifferKRCG21}, or the low-rank matrices in the query and value projection of the self-attention networks \cite{HuSWALWWC22}.
However, limited works examine the impact of sub-network structure and integrate   pruning methods with PETL methods.
In fact, studies in network pruning have shown that the modeling ability of neural networks relies not only on the parameters but also on the sub-network structures \textcolor{black}{that are decided by the pruning masks}. 
For instance, \citet{ZhouLLY19} discovered that the sub-network of an untrained model can yield great performance without any parameter update. In light of this, we seek to incorporate the structural information of sub-networks into  PETL. We believe that when enough structural information is injected into the network, we no longer need that many parameters in PELT modules and further improve the parameter efficiency.
 

To this end, we propose a novel PETL method dubbed \textbf{\propet} that enables efficient sharing of a single \textit{prototype} adapter, prefix, or LoRA across layers and tasks. When sharing the prototype network, \propet learns binary masks to prune different sub-networks in different layers and tasks (Figure~\ref{fig:shareadapter}).
The connections of the prototype network pruned in one layer can be used in another with a different pruning mask.
In this way, a parameter can be used multiple times across different modules, achieving higher parameter efficiency. 
Previous methods~\cite{abs-2210-04284, abs-2210-04457} only consider simply discarding (pruning) the useless parameters, while we focus on the structural information in masks by strategically dispatching the parameters in the single prototype network to different modules.
We evaluate \propet on various downstream tasks, including GLUE~\cite{wang-etal-2018-glue}, XSum~\cite{NarayanCL18}, and WMT16 Ro-En~\cite{wmt16}.
Experiments show that \propet achieves better performance than other PETL methods while using significantly fewer parameters.

Our contributions are summarized as follows:

 \begin{itemize}
  \item
We propose \propet, a highly parameter-efficient transfer learning method that injects structural information into PELT and allows for efficient sharing of a single prototype network across layers and tasks.
  \item
   
Experiments show \propet is able to dramatically reduce the storage of the parameters while achieving better performance than conventional PETL methods.
\item
\propet offers an alternative view for network pruning and sharing, 
where we use binary masks to decide when to discard or share the parameters. We hope to inspire more intriguing explorations in this direction.

\end{itemize}

\section{Related Work}
In this section, we briefly survey ideas that are related to our work from three fields: parameter-efficient transfer learning, pruning methods, and multi-task learning.

\subsection{Parameter-Efficient Transfer Learning}
Recently, as the pre-trained language models get larger and larger, some parameter-efficient transfer learning methods that only update a few extra parameters while freezing the PLM backbone have been proposed. 
Adapter-tuning~\cite{HoulsbyGJMLGAG19} fine-tuned adapter modules inserted after each attention and feed-forward layer. 
Prefix-tuning~\cite{LiL20} placed an additional trainable prefix to the keys and values matrix in the attention module. 
LoRA~\cite{HuSWALWWC22} injected tunable rank decomposition matrices into each Transformer layer.
Based on these parameter-efficient transfer learning methods, \citet{HeZMBN22} gave a unified framework that allows for the transfer of design elements across various PETL approaches.
However, when applied to larger PLMs and a broader range of tasks, these methods still require a large storage space because the number of extra parameters is directly proportional to the number of layers and tasks.
Inspired by the parameter-sharing techniques of ALBERT~\cite{LanCGGSS20}, we propose sharing the additional parameters in PETL modules across layers and tasks.
Our method can thus obtain higher parameter-efficient efficiency with a significantly smaller portion of additional  storage than existing PETL methods.

\subsection{Pruning Methods}

Pruning is one of the most popular methods to reduce unnecessary weights from over-parameterized neural networks while maintaining comparable performance.
Recently, \citet{FrankleC19} proposed Lottery Ticket Hypothesis and stated that in a randomly initialized dense model, a sparse sub-network exists that, when trained in isolation, can achieve performance comparable to dense models.
Accompanied by this hypothesis, many pruning-before-training methods have emerged \cite{LeeAT19, bai2022parameterefficient,sreenivasan2022rare}.  \textcolor{black}{\citet{xu-etal-2021-raise} further propose a method that prunes the backward gradient of the neural network, as opposed to pruning the network parameters themselves.}
Based on these methods, some works~\cite{abs-2210-04284, abs-2210-04457} also proposed to combine pruning algorithms with parameter-efficient methods to further decrease the additional module size. 
However, those methods only focus on discarding redundant parameters. A parameter is either discarded or retained without any sharing.
They fail to make full use of the additional parameters and cannot achieve highly sparse sub-networks without significantly compromising accuracy.



\begin{figure*}[ht]
    \centering
    \includegraphics[width=1.01\textwidth]{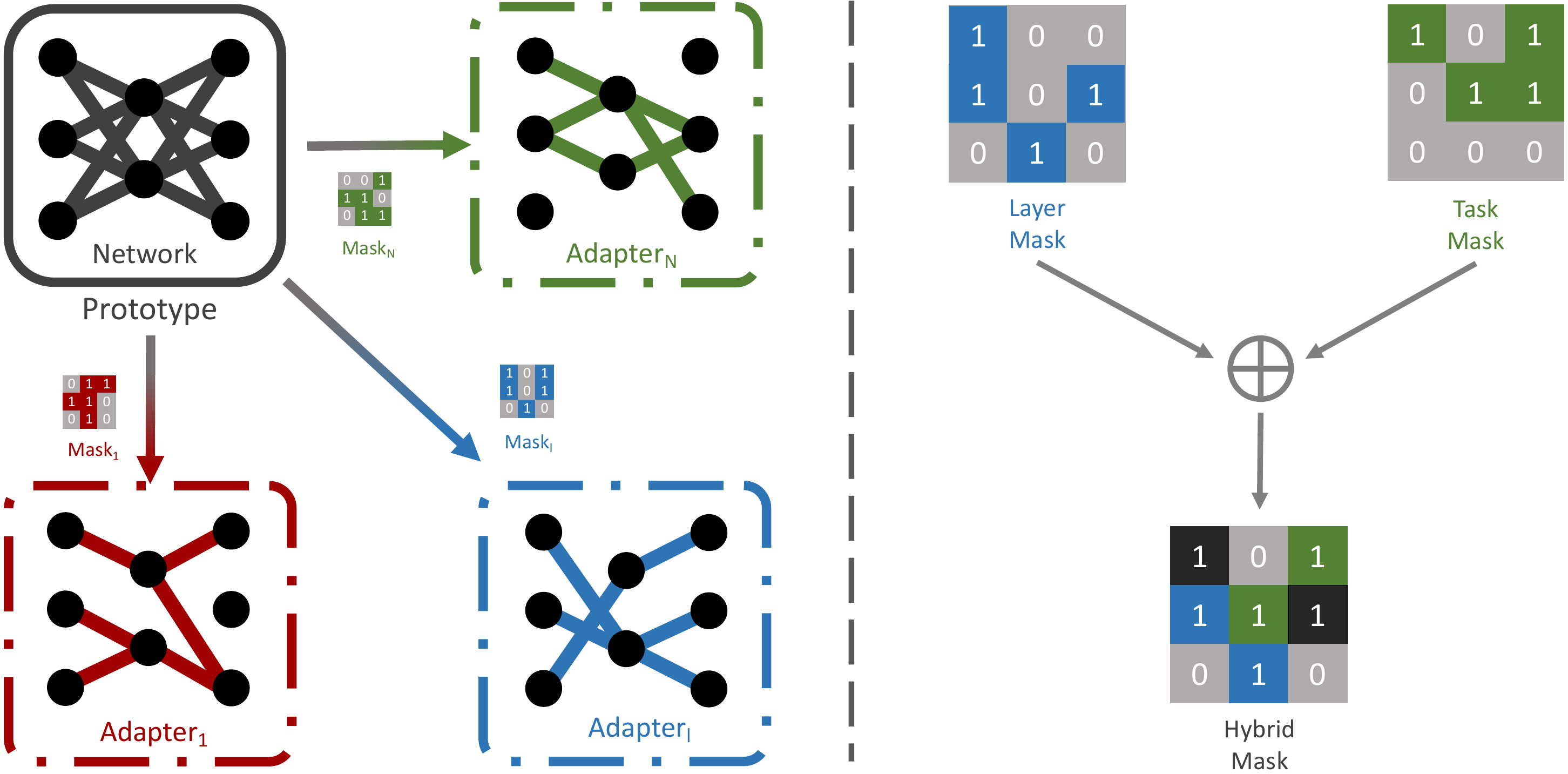}
    \caption{Overview of our \propet method. In the left part, the grey neural network indicates the prototype network. Using various binary masks, we can derive sub-networks by pruning certain connections, as depicted by the green (top right), red (down left), and blue (down right) networks. In the right part, we learn layer masks and task masks under multi-task learning. Given a specific Transformer~\cite{NIPS2017_3f5ee243} layer and task to handle, \propet generates a hybrid mask by performing an OR logical operation on the layer mask and the task mask. It then uses hybrid masks to generate different sub-networks from the prototype.}

    \label{fig:propet}
\end{figure*}

\subsection{Multi-Task Learning}
Multi-task learning~\cite{ZhangY22}, which involves training a single model to perform well on multiple tasks, has gained popularity as a research direction in machine learning.
However, this approach can be hindered by catastrophic forgetting~\cite{KirkpatrickPRVD16} and data imbalance among tasks, which will result in overfitting on low-resource tasks and underfitting on high-resource tasks~\cite{abs-1907-05019}.
\citet{HoulsbyGJMLGAG19} propose adapter tuning that only introduces and updates small additional parameters for each task while freezing the pre-trained model.
Based on such a parameter-efficient method, \citet{MahabadiR0H20} train a hyper-network named Hyperformer, which generates task-specific weights for the adapter modules when fed with different task embeddings.

\section{Methods}
In this section, we first give an introduction to parameter-efficient transfer learning (PETL). We then present our method \propet, depicted in Figure~\ref{fig:propet}, which combines the techniques of parameter-sharing and pruning to further improve the parameter efficiency compared to existing PETL methods.
\label{preliminary}
\vspace{-1mm}
\subsection{Preliminaries}
In parameter-efficient transfer learning, we freeze the parameters $\theta_{lm}$ of the pre-trained language model and then introduce additional fine-tunable parameters denoted as $ \theta_t $.
Given dataset $\{X_i, Y_i\}^N_{i=1}$, the goal of parameter-efficient fine-tuning is to maximize the following likelihood of the label $Y$ by only updating the additional parameters $\theta_t$:
\begin{equation}\label{eq:eq1}
\operatorname*{max}_{\theta_t}  \sum_{i=1}^N  \log P(Y_i | X_i; \theta_{lm}, \theta_t)
\end{equation}

Such parameter-efficient methods suggest a more effective way to adapt pre-trained language models over downstream tasks than fully fine-tuning. We give a brief introduction of the three most widely used PETL modules, namely adapter~\cite{HoulsbyGJMLGAG19}, prefix-tuning~\cite{LiL20}, and LoRA ~\cite{HuSWALWWC22} in Appendix \ref{app:PETL module}.
However, there is still a storage limitation when we handle a large range of tasks using these methods.
\textcolor{black}{In this paper, we investigate the potential for further enhancing parameter efficiency in neural network models by reducing storage requirements. While previous PETL methods have primarily focused on decreasing the number of parameters to improve efficiency, our approach posits that employing varying bit lengths (e.g., 1-bit, 8-bit, 32-bit) during storage can lead to significant improvements in parameter efficiency by reducing the overall number of bits used by the parameters.}
\textcolor{black}{To this end, we use bits to measure the storage, which we call Bit-Level Storage (BLS), to take into account the fact that different parameters may have different bit lengths.
Consider a neural model, where each parameter has a specific bit length. Then we divide these parameters into $N$ distinct groups based on their respective bit lengths. Let $\{\rho_i\}_{i=1}^N$ denote the number of parameters within the group $i$, with corresponding bit lengths $\{b_i\}_{i=1}^N$. The BLS for these parameters can subsequently be determined as follows:}

\begin{equation}\label{eq:eq1}
\text{Bit-Level Storage} = \sum_{i=1}^N \rho_ib_i
\end{equation}


\subsection{Shared Prototype Network}
Parameter-efficient methods like adapter and prefix-tuning tend to introduce an additional module in each Transformer layer.
Assuming the Transformer has $L$ layers, we can split the parameters $\theta_t$ into $[\theta_{t,1}, ...,\theta_{t, L}]$ according to their layer indexes.
Therefore, we can rewrite Equation~\ref{eq:eq1} as:
\begin{equation}\label{eq:eq2}
\operatorname*{max}_{\theta_{t,1}, ... ,\theta_{t,L}}  \sum_{i=0}^N  \log P(Y_i | X_i; \theta_{lm}, [\theta_{t,1}, ... ,\theta_{t,L}])
\end{equation}

Inspired by ALBERT \cite{albert}, in our methods, we first introduce additional parameters for a single PETL module as our prototype network, denoted as $\theta_{pro}$.
Then, we share the prototype network across different layers.
Assuming that the number of parameters in a single PETL module is $n$, we can decrease total parameters from $nL$ to only $L$, which significantly improves the parameter efficiency.
Therefore, we convert the objective function to a more concise one:
\begin{equation}\label{eq:share}
\operatorname*{max}_{\theta_{pro}}  \sum_{i=0}^N  \log P(Y_i | X_i; \theta_{lm}, \theta_{pro})
\end{equation}


\begin{algorithm}[t!]\small
\caption{\propet training algorithm}\label{alg:propet}
\SetKwInput{KwInput}{Input}                
\SetKwInput{KwOutput}{Output}              
\DontPrintSemicolon
  
  \KwInput{Dataset $\mathcal{D}= {(x_i, y_i)}$, prototype network learning rate $\lambda_p$, mask learning rate $\lambda_m$, 
  number of layers $L$, sparsity $k\% \in [0,1]$, pretrained parameters $\theta_{lm}$}
  \KwOutput{Parameter of the prototype network $\theta_{pro} \in\mathbb{R}^n$, binary mask  across layers $m \in\mathbb{R}^n$}
        $\theta_{pro} \gets$ randomly initialize in $\mathbb{R}^n$\;
        $s \gets$ randomly initialize in $\mathbb{R}^n$\;
        
        \For{$ (x_i, y_i) \; in\; D$}
        {
            \Comment{Apply masks in each layer}
            \For{$ l \; in\; 1, 2,..., L$}
            {
                $m_l\gets h(s_l, k)$ \\
                $\theta_{sub, l} \gets \theta_{pro} \odot m_l$
            }
            $\theta_{pro} \gets \theta_{pro}-  \lambda_p \nabla_{\theta_{pro}} \log P(y_i | x_i; \theta_{lm}, \theta_{sub})$ \\
            $s \gets s - \lambda_m \nabla_{s} \log P(y_i | x_i; \theta_{lm}, \theta_{sub})$ \;
        }
         \For{$ l \; in\; 1, 2,..., L$}
         {
         $m_l\gets h(s_l, k)$ 
         }
        \KwRet $\theta_{pro}, m $\;

\end{algorithm}

\subsection{Masked Sub-Networks}\label{sub:supermask}

Sharing the parameters alone will reduce the model's capacity to capture meaningful representation in different layers, leading to suboptimal results.
Inspired by \citet{ZhouLLY19} and \citet{RamanujanWKFR20}, we believe that parameters and network structures are both crucial contributing factors to the model's representative capacity.
To this end, we introduce different binary masks $m_l = \{0, 1\}^n$ in each Transformer layer $l$ (left part of Figure \ref{fig:propet}), where $n$ denotes the number of parameters in a single PETL module.
Each mask represents a corresponding subnetwork of the shared prototype network.
Even though the prototype is shared among all layers, we can use different masks to create different sub-networks for each layer $l$, whose parameter will be $\theta_{sub, l} = \theta_{pro} \odot m_l$, where $\odot$ indicates the element-wise product.
With this, we can get our final objective function as:
\begin{equation}\label{eq:final}
\begin{aligned}
\operatorname*{max}_{\theta_{pro}, m_1, m_2, ..., m_L}  \sum_{i=0}^N  \log P(Y_i | X_i; \theta_{lm}, \theta_{sub}) 
\end{aligned}
\end{equation}

where $\theta_{sub}$ = [$\theta_{pro} \odot m_1$, $\theta_{pro} \odot m_2$, ..., $\theta_{pro} \odot m_L$].

To learn such masks, we develop our training algorithm based on the edge-popup approach~\cite{RamanujanWKFR20}.
Specifically, for each binary mask $m_l$, we introduce floating-point scores $s_l \in\mathbb{R}^n $.
In the forward pass, we generate the binary mask $m_{l}$ by setting the top-$k$\% with the largest absolute value in $s_{l}$ as 1 and the rest as 0. We denote such top-$k$\% thresholding function as $h$, and  $m_{l} = h(s_l)$. We refer to the hyperparameter $k\%$ as the sparsity ratio in subsequent sections of this paper. 
During the backpropagation, we use the straight-through gradient estimator~\cite{BengioLC13} to approximate the gradient of the scores, where function $h(\cdot)$ is treated as the identity function.
In addition to training the masks, we also jointly optimize the prototype network.

\textcolor{black}{Our approach employs learnable floating-point scores for each binary mask during fine-tuning, leading to $nL$ parameters. When integrated with the prototype network, \propet updates a total of $nL+n$ parameters during the training phase, which is marginally more than the conventional PEFT methods which have $nL$ extra parameters. A comprehensive overview of the \propet algorithm can be found in Algorithm~\ref{alg:propet}.
After training, we discard the floating-point scores and retain only the binary masks (1-bit) together with the shared prototype network (32-bit).
Assuming that the 32-bit prototype network requires $p$ bit-level storage and the binary mask of the same dimension demands $p/32$, our \propet can achieve a substantial decrease in storage from around $pL$ to $p(1+L/32)$. To reconstruct the network structure during inference, we adhere to the following steps: (1) first load the unpruned, shared 32-bit prototype network, (2) then load the binary masks (1-bit) for each layer/task, and (3) extract and use the pruned subnets from the shared prototype network based on specific binary masks during the inference step.}


\subsection{Hybrid Masks for Multi-Task Learning}
Rather than just sharing a PETL module across layers under single-task learning, we can also allow for efficient sharing of the prototype network across multiple tasks.
In our approach, we leverage layer masks, as introduced in the previous section, to support parameter sharing within the model. Additionally, we introduce task masks to support parameter sharing across multiple tasks.
By performing a logical OR operation\footnote{We provide ablations regarding the choice of mask combining methods in Appendix \ref{mask_combine_ablat}.} on these masks, we can obtain a hybrid mask for a specific layer in a specific task, as shown in the right side of Figure~\ref{fig:propet}. 
\begin{equation}\label{eq:mask}
    m_{hybrid} = m_{layer} \lor m_{task}
\end{equation}

With the design of the hybrid mask, given $T$ tasks and $N$ layers in the pre-trained language models, we only require one \propet module, $N$ layer masks, and $T$ task masks, further reducing the BLS from the single task scenario (e.g., 0.011\% BLS as in Table \ref{tab:multitask_GLUE}). 
In addition, layer masks and task masks, which can be combined as hybrid masks, will potentially help infuse the knowledge into the shared prototype network from layers and tasks.

\section{Experimental Setup} \label{experimental setup}
We briefly summarize the experimental setup in this section. More details can be found in Appendix \ref{app:experiment details}.

\paragraph{Datasets}
We evaluate \propet on a wide range of benchmarks, including language understanding (GLUE~\cite{wang-etal-2018-glue}), text summarization (XSum~\cite{NarayanCL18}), and machine translation (WMT16 Ro-En~\cite{wmt16}). 

\paragraph{Backbones}
We use \robertabase~\cite{roberta} for single-task learning on GLUE. During fine-tuning, we only tune our \propet module and the text classification head. For generation and multi-task learning benchmark, we use \basebase~\cite{2020t5} and only tune the \propet module. Note that some previous works also tune the layer norms \cite{HoulsbyGJMLGAG19, MahabadiR0H20} while we keep them frozen during fine-tuning.

\paragraph{PETL Modules and \propet}

We use the Pfeiffer adapter~\cite{PfeifferKRCG21} as our adapter module and set the bottleneck dimension as 64 by default.
For prefix-tuning, we follow \citet{LiL20} and choose the prefix length to be 64.
As for LoRA tuning~\cite{HuSWALWWC22}, the bottleneck dimension and the scaling factor $\alpha$ are both set to 32. 
In \propet, we increase the value of $\alpha$ to 48 to scale the representation, as the sparse network will decrease the norm of the output representation.
We give a brief summary of these PETL modules and how \propet is implemented on top of them in Appendix \ref{app:PETL module}.
Following~\citet{RamanujanWKFR20}, we choose the sparsity ratio $k$\% of \propet as 0.5, which we will also further discuss in Section~\ref{sparsity}.
In multi-task learning, we aim to maintain an expected $k$\% around 0.5 for the hybrid mask, so we set the $k$\% to 0.3  for both the layer and task masks.\footnote{Two random and independent binary masks whose elements have 30\% probability to be one will produce a resulting mask with about 50\% ones after the OR logical operation, which can be calculated using the equation $P(A \cup  B) =P(A) + P(B) - P(A)P(B)$. }

\begin{table*}[htp]
\small
\setlength\tabcolsep{3.4pt}
\centering
{
\scalebox{0.9}{
\begin{tabular}{l |r r |  c c c c c c c c|c }
\toprule
Model &\textcolor{black}{\% FT Params}  & \textcolor{black}{\%BLS} & CoLA & SST-2 & MRPC & QQP & STS-B & MNLI & QNLI & RTE & Avg \\
\midrule
\robertabase & \textcolor{black}{100.00\%} &100.00\%& 60.94 & 94.04 & 87.25/90.76 & \textbf{91.34}/\textbf{88.53} & 90.96/90.70 & \textbf{87.57} & 92.53 & 73.14 &  86.16\\
\midrule

Prefix ($l$=64) &  \textcolor{black}{0.95\%} & \textcolor{white}{00}0.95\%&63.27 & \textbf{94.42} & \textbf{89.05}/\textbf{92.01} & 88.86/85.18 & 90.46/90.39 & 85.76 & 91.46 & 63.79 &  84.97\\
LoRA (bn=32) &  \textcolor{black}{0.95\%} & \textcolor{white}{00}0.95\%&62.24 & 93.81 & 86.76/90.48 & 88.79/85.15 & 90.73/90.49 & 86.59 & 91.85 & 67.63 &  84.96 \\
Adapter (bn=64) &  \textcolor{black}{0.95\%} & \textcolor{white}{00}0.95\%  & 63.15 & 94.00 & 86.93/90.49 & 89.78/86.52 & 90.84/90.65 & 87.10 & 92.23 & 70.50 &  85.65\\

\midrule
\propetprefix ($l$=64) &  \textcolor{black}{1.03\%} & \textcolor{white}{00}0.11\% & 61.81 & 94.00 & 87.42/91.00 & 88.85/85.22 & 90.48/90.47 & 85.73 & 91.05 & 63.79 &  84.53\\

\propetlora (bn=32) &  \textcolor{black}{1.04\%} & \textcolor{white}{00}0.11\%& 62.16 & 93.62 & 88.73/91.80 & 87.59/83.71 & 90.92/90.83 & 85.30 & 91.75 & 72.66 &  85.37\\

\propetadapter (bn=64) &  \textcolor{black}{1.04\%} & \textcolor{white}{00}0.11\%& \textbf{65.43} & 94.15 & 88.24/91.41 & 89.40/86.04 & \textbf{91.34}/\textbf{90.95} & 86.53 & \textbf{92.58} & \textbf{76.50} &  \textbf{86.60}\\
\bottomrule
\end{tabular}
}
}
\caption{Performance of all models based on RoBERTa on the GLUE tasks under single task settings. Bold fonts indicate the best results. ``bn'' stands for the bottleneck dimension and `` $l$'' refers to the number of prefixes. 
\textcolor{black}{Here \textit{\% FT Params} refers to the percentage of fine-tunable parameters during training (including the underlying floating point score of each pruning mask). \textit{\%BLS} indicates the task-specific Bit-Level Storage (defined in Sec \ref{preliminary}) calculated against the fully-finetuned counterparts when saving the model weights and during the inference time.} 
\vspace{-1em}
}
\label{tab:single}

\end{table*}

\paragraph{Evaluation} For text generation, we report ROUGE-2~\cite{lin-2004-rouge} on the XSUM test set  and BLEU~\cite{bleu} score on the Ro-En test set. Since the test sets of GLUE are not released publicly, following \citet{zhang2021revisiting} and \citet{mao-etal-2022-unipelt}, when the sample number of the datasets is fewer than 10k (RTE, MRPC, STS-B, CoLA), we divide the original validation set into halves -- the first half for validation and the second for testing.
As for the other datasets in GLUE, we randomly choose 1k samples from the training set as our validation data and test on the original validation set. we report both accuracy and F1 for MRPC and QQP in  GLUE . For STS-B, we report both Pearson and Spearman correlation coefficients. For CoLA, we report Matthews correlation. For all remaining sub-tasks in GLUE, we report accuracy. 
Due to high training overhead for generation tasks, we report experimental results with one run for XSum and Ro-En.
For GLUE, we report the mean of three different random runs.



\section{Results}
\subsection{Single-Task Learning}
\paragraph{Results in Language Understanding} 


In Table \ref{tab:single}, we report the performance of \propet and various baselines on the GLUE benchmark. \textcolor{black}{Both \propetadapter and \propetlora demonstrate superior performance compared to their respective counterparts (adapter and LoRA). Despite having slightly more parameters during the fine-tuning process, \propet requires only 1/9 (0.11\% v.s. 0.95\%) of bit-level storage during inference, making them more efficient options.}
Specifically, \propet increases the average score of the adapter by 0.95 and improves the score of LoRA by 0.41. Besides, \propetadapter remarkably outperforms the fully fine-tuned model (86.60 v.s. 86.16) while using $0.11\%$ storage of the fully fine-tuned model.
These results indicate that although with reduced parameters, \propet injected with the structure information from masks can make more use of the single prototype network and achieve better performance compared to their counterparts.
However, we also found that \propetprefix did not outperform prefix-tuning, which we believe is caused by the reparameterization in prefix-tuning that has a harmful effect on the mask learning.\footnote{See more details and explanation in Appendix~\ref{app:prefix-tuning}}
Overall,  \propet increases the performance of the adapter to the greatest extent. \propetadapter also achieves the highest performance among the three \propet variants.
We will stick to \propetadapter for the rest of the experiments.


\paragraph{Results in Language Generation}
To verify \propet can also be applied to harder tasks,  we evaluate our method on two language generation datasets, XSum and WMT16 Ro-En. The results are presented in Figure \ref{dist} (a) and (b).
We find that \propetadapter can perform just as well as the regular adapter method while using significantly less \textcolor{black}{bit-level storage}. Additionally, when consuming more than 1.6\% of the \textcolor{black}{storage},  \propetadapter is able to achieve competitive performance on the XSum dataset compared with the fully fine-tuned T5. However, both adapter tuning and \propetadapter do not reach the level of the fully fine-tuned model on Ro-En. One potential explanation is Ro-En is harder because translation knowledge may not be covered a lot during the pre-training process of T5. To perform well on such tasks, the model needs to learn additional knowledge and requires more tunable parameters during fine-tuning. We note that such sub-optimal performance on hard generation tasks is not only a nature of \propet but generally exists in all PETL methods. Similar findings are also presented in ~\citet{2020t5} and ~\citet{HeZMBN22}.
Overall, these experiments show that our \propet is also more parameter-efficient on text generation benchmarks compared to existing PETL methods.



\subsection{Multi-Task Learning}
\begin{table*}[htbp]
\small
\setlength\tabcolsep{3.6pt}
\centering
{
\scalebox{0.87}{
\begin{tabular}{l | r r | c  c c c c c c c | c }
\toprule
Model & \pbox{3cm}{\color{black}{\%FT Params} \\ \raggedleft per task\vspace{0.1em}}   & \pbox{3cm}{\color{black}{\%BLS} \\ \raggedleft per task\vspace{0.1em}} & CoLA & SST-2 & MRPC & QQP & STS-B & MNLI & QNLI & RTE & Avg \\
\toprule
\rowcolor{white!20}\multicolumn{11}{l}{\it \textbf{Single-Task Learning}}\\
\midrule
\basebase\textsuperscript{\textdagger} & \textcolor{black}{100.0\%} & 100.000\% & 54.85 & 92.19 & 88.18/91.61 & 91.46/88.61 &89.55/89.41 & 86.49 & 91.60 & 67.39 & 84.67\\
Adapter (bn=64) & \textcolor{black}{1.070\%} &1.070\% & 62.64 & 94.07 & 87.36/91.06 &90.25/87.28 &89.88/89.55 &85.76 &92.85 & 71.01&85.61\\
\midrule
\rowcolor{white!20}\multicolumn{11}{l}{\it \textbf{Multi-Task Hypernetworks}}\\
\midrule
Hyperformer++\textsuperscript{\textdagger} & \textcolor{black}{0.290\%} & 0.290\% &    63.73 &        94.03 &        89.66/92.63 &       90.28/87.20 &               90.00/89.66 &        85.74 &        93.02 &       75.36 & 86.48  \\ 
\midrule
\rowcolor{white!20}\multicolumn{11}{l}{\it \textbf{Multi-Task Training}}\\
\midrule

\basebase\textsuperscript{\textdagger} &  \textcolor{black}{12.500\%} &  12.500\% &         54.88 &        92.54 &        \textbf{90.15}/\textbf{93.01} &      \textbf{91.13}/\textbf{88.07} &            88.84/88.53 &        \textbf{85.66} &        92.04 &       75.36 &  85.47\\
Adapter (bn=64) & \textcolor{black}{0.130\%} & 0.130\%& \textbf{62.08} & 93.57 & 89.49/92.64 &90.25/87.13 & 87.54/87.41 & 85.14 &92.80  & 72.22 & 85.48 \\
Adapter (bn=6) & \textcolor{black}{0.013\%} & 0.013\%& 58.34 & 93.61 & 86.20/90.44 &90.10/86.98 & 86.96/86.66 &84.02 &92.38& 67.63 &83.94\\
\propetadapter (bn=64) & \textcolor{black}{0.156\%} & 0.011\%& 61.43 & \textbf{94.22}  & 87.36/90.97 & 90.13/87.14 &90.32/90.12 &85.34  & \textbf{93.01}  & 75.60  & \textbf{85.97} \\
\propetadapter (bn=6)  & \textcolor{black}{0.016\%} & \textbf{0.001}\%&54.59 & 93.53  & 87.36/91.02 & 90.15/87.04 &\textbf{90.70}/\textbf{90.50} &85.08  & 92.79  & \textbf{75.86}  & 85.32 \\
\bottomrule

\end{tabular}
}
}
\caption{GLUE Results on T5. Under single-task learning, we train each task with different model copies. As for multi-task training, we train a unified model or adapter. Results marked with \textdagger\xspace are from the implementation of \citet{MahabadiR0H20}. Bold fonts suggest the best results in the block.
}
\label{tab:multitask_GLUE}

\end{table*}
We present the results of multi-task learning and also provide baseline results from single-task learning using the \basebase in Table~\ref{tab:multitask_GLUE}.
Our best-performing model, \propetadapter with bottleneck dimension of 64, surpasses the fully fine-tuned T5. We also compare \propet with Hyperformer++ ~\cite{MahabadiR0H20}, a hyper-network that is specifically designed to transfer knowledge across tasks. The latter uses significantly more \textcolor{black}{task-specific bit-level storage} (26$\times$: $0.011\%$  v.s.  $0.29\%$ per task), while only increasing the average score by $0.51$.
\textcolor{black}{Compared to the vanilla adapter, \propetadapter can  be marginally better under a similar fine-tuning parameter budget but  with 1/9 of the original storage.}
Besides, we experiment with an extreme case, where we set the bottleneck dimension to 6. Our results show that the accuracy of adapter tuning decreases from 85.48 to 83.94, while \propetadapter still maintains a comparable performance to the fully fine-tuned model (85.32 v.s. 85.47) with a remarkably small percentage ($0.001\%$) of \textcolor{black}{bit-level storage} per task.
This demonstrates that normal adapter tuning can not make full use of the parameters and may fail to perform well with a relatively small bottleneck dimension.
However, in the case of \propetadapter, even with a bottleneck dimension that is only 6, it can still achieve a reasonable result. 
\textcolor{black}{To further validate that \propet is effective for larger-sized models, we also carry out experiments on the T5 3B variant and present the findings in Table \ref{T53B}. The outcomes align with our conclusions drawn from the T5 base model.}
We believe that, in \propetadapter, the structure information learned by the mask can make up for the performance drop due to the shrink of the bottleneck dimension to a certain extent.
We further compare adapter tuning with \propetadapter with different percentages of \textcolor{black}{bit-level storage} by varying the bottleneck dimension.
\begin{table}[t!]
\small
\setlength\tabcolsep{6.1pt}
\centering
\scalebox{0.9}{
\begin{tabular}{l  r  c  c}
\toprule
 Model & \pbox{3cm}{\color{black}{\%FT Params} \\ \raggedleft per task} & \pbox{3cm}{\color{black}{\%BLS} \\ \raggedleft per task\vspace{0.1em}} &  Avg\\ 
\midrule

\basexlarge &  \textcolor{white}{0.00}25\% & \textcolor{white}{0.00}25\%  & 88.92  \\
Adapter (bn=64) &  0.0278\% & 0.0278\%  & 88.31  \\
\propetadapter (bn=64) &  0.0283\% & 0.0016\%  & \textbf{89.02}  \\

\bottomrule
\end{tabular}
}
\caption{Multi-task learning results on the GLUE using \basexlarge as the backbone.}
\vspace{-1em}
 \label{T53B}
\end{table}
The results are presented in Figure \ref{dist} (c). It shows \propetadapter is able to reach the fine-tuned performance with as few as $0.002\% $ \textcolor{black}{task-specific bit-level storage}. We can also see that the curve shares a similar trend with those in Figure \ref{dist} (a) and (b), which we will further discuss in the next section. \looseness=-1

\section{Discussion}
\begin{figure*}[t!] 
	\centering
	\begin{subfigure}{0.31\linewidth} 
		\centering
		\includegraphics[width=1.0\linewidth]{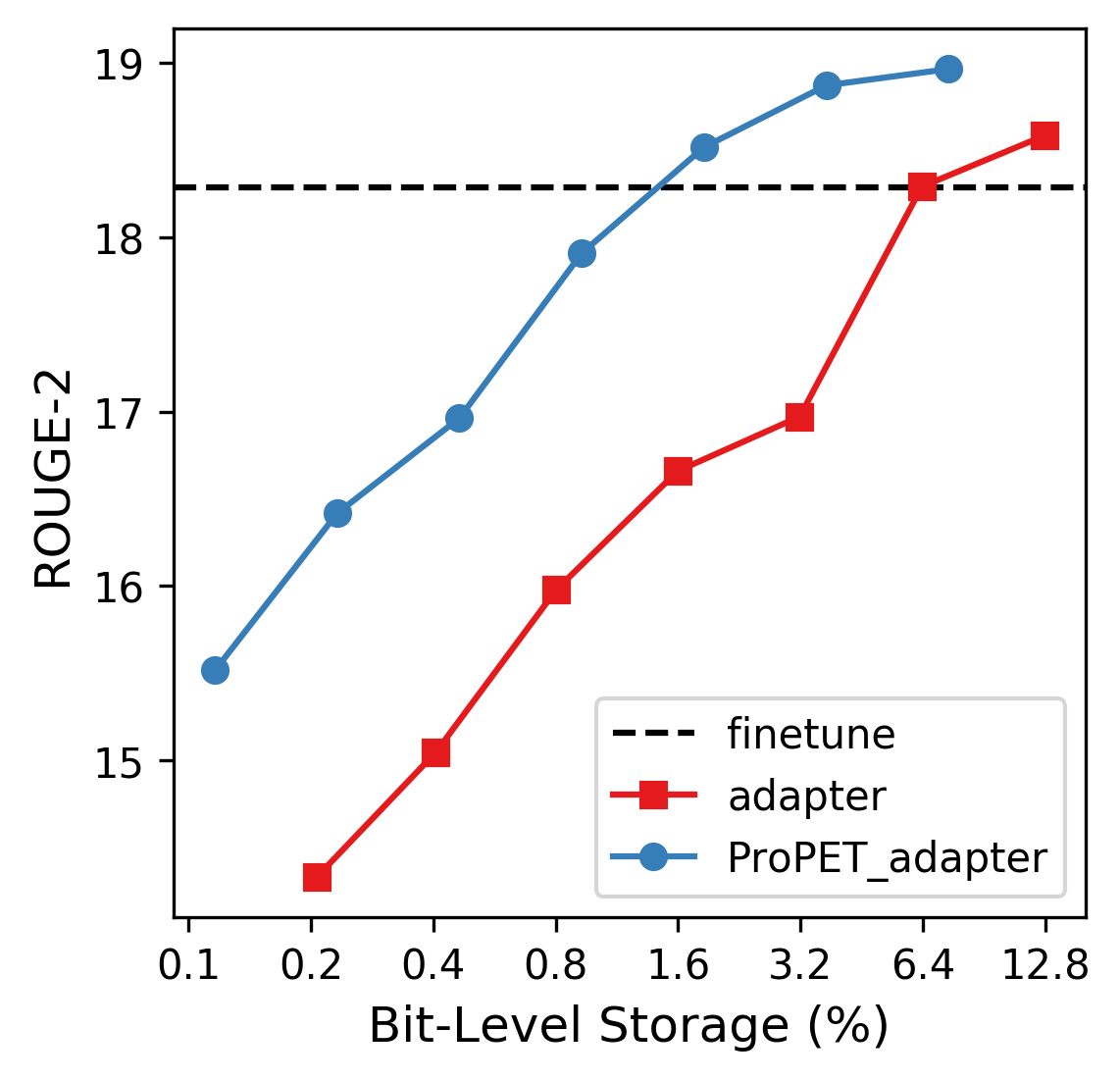}
		\caption{XSum}
	\end{subfigure}
	\centering
	\begin{subfigure}{0.31\linewidth}
		\centering
		\includegraphics[width=1.0\linewidth]{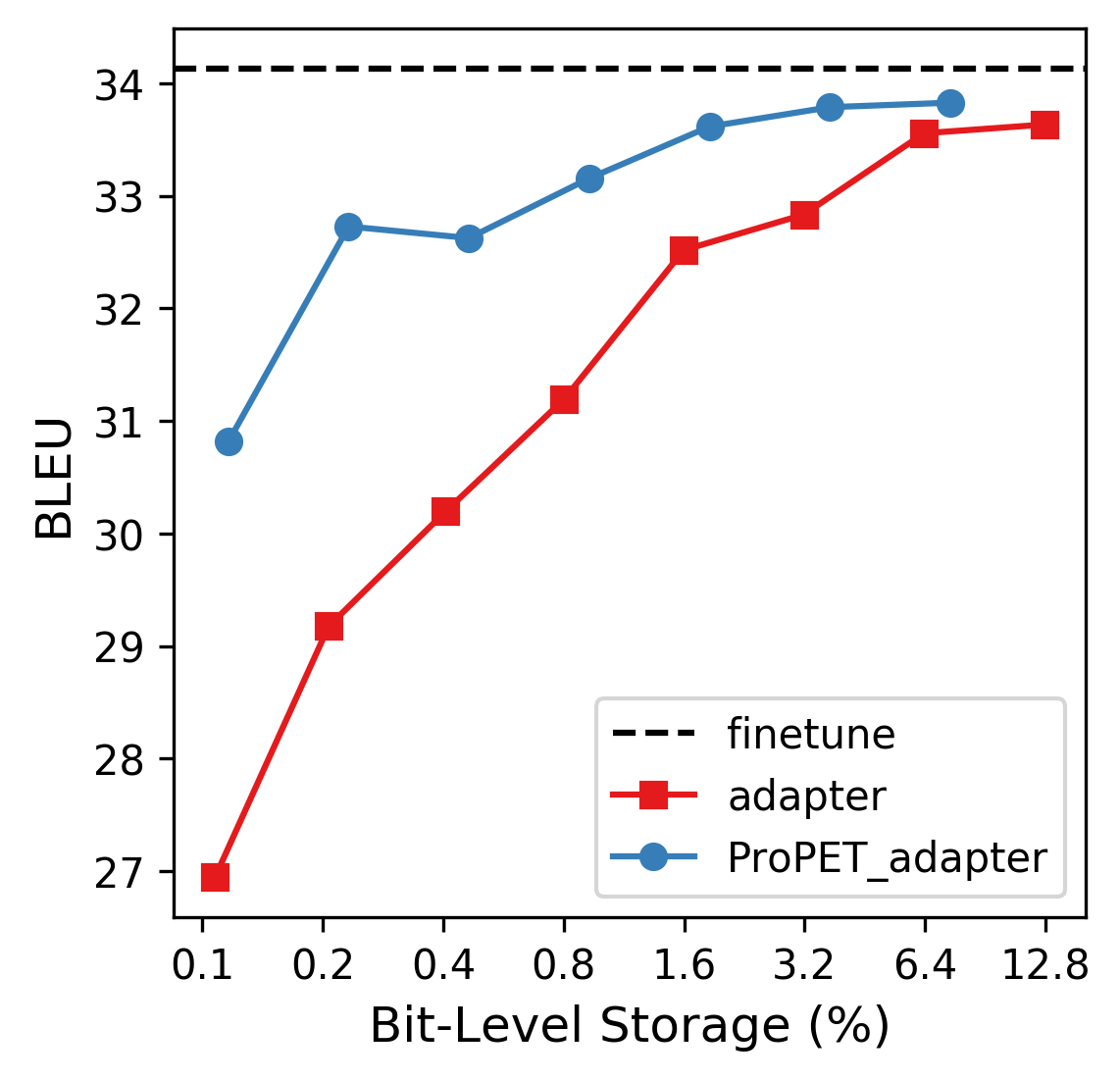}
		\caption{Ro-En}
	\end{subfigure}
	\centering
	\begin{subfigure}{0.315\linewidth}
		\centering
		\includegraphics[width=1.0\linewidth]{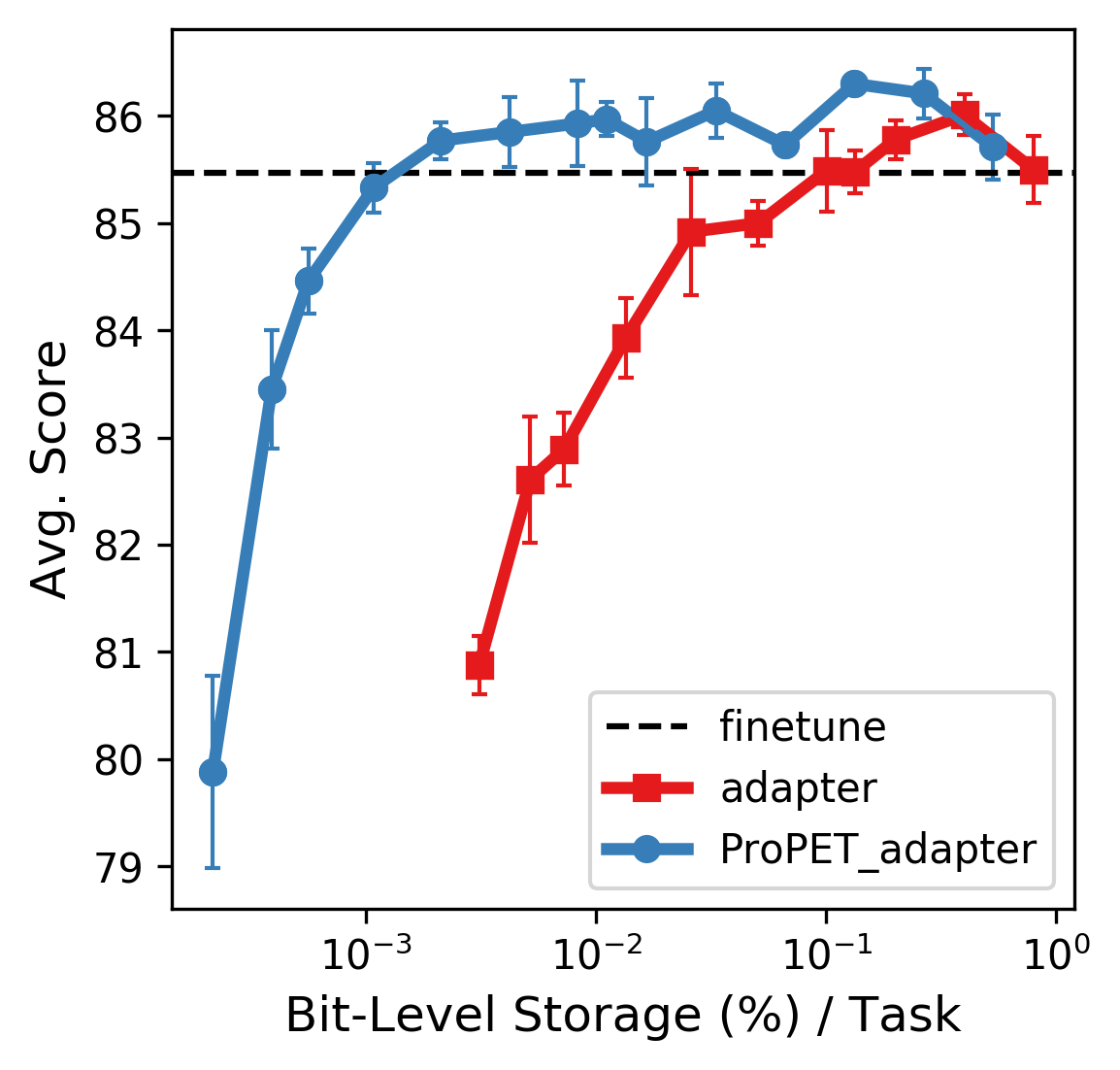}
		\caption{GLUE}
	\end{subfigure}
	\caption{Performance of adapter and \propetadapter on XSum (left), Ro-En (middle), and GLUE (right) with \basebase model. We train on XSum and Ro-En under single-task settings for 1 run. We train GLUE under multi-task learning and report the average score with 3 runs. We additionally provide these results in table format in Appendix~\ref{hugetable}.}
	\label{dist}

\end{figure*}
\paragraph{How does \propet Scale Differently to Adapter?}
Figure \ref{dist} presents the results when we adjust the size of the adapter and \propetadapter on three different datasets.
Despite the difference in tasks and methods, we discover that adapter and \propetadapter show similar scaling trends on all three datasets when we increase the proportion of the \textcolor{black}{task-specific bit-level storage}. Their performance increases linearly with respect to the log scale of the extra storage  in the beginning. When the adapter and \propetadapter reach close to the performance of the fully fine-tuned model, their performance gradually converges.
Even though \propetadapter and adapter tuning can slightly exceed the fully fine-tuned performance on some datasets, their performance is still bounded to the same level and cannot outperform the fully fine-tuned model by a large margin. For instance, the performance of both adapter and \propet starts to drop when the \textcolor{black}{task-specific storage} exceed 0.4\% per task  on GLUE (Figure \ref{dist} (c)). 
However,  \propetadapter is able to reach the fully fine-tuned level much earlier in the scaling curves. \textcolor{black}{Given a fixed amount of task-specific storage}, \propetadapter is also able to achieve better results than the adapter. These results indicate that our share-and-mask method is overall more  efficient than the adapter across almost all scales.

\paragraph{Which is More Important, Sharing or Masking?}
In this section,  we discuss the effect of masking and sharing in the prototype network by comparing \propet with \textcolor{black}{a random mask baseline} and two alternative settings (\textit{only mask} and \textit{only share}). \textcolor{black}{For the random mask setting, we randomly select the sub-network of the prototype module during the forward process rather than relying on the updated mask score.} \textit{Only mask} involves not sharing the module across layers but only learning masks to prune different PETL modules into sparse sub-networks. \textit{Only share} shares a single PETL module among layers without using masks for pruning. 
Masking without sharing will drastically increase the number of fine-tunable parameters if we keep the same bottleneck dimension and sparsity ratio.
To keep the number of fine-tunable parameters on the same level, we either keep the bottleneck dimension the same and reduce the sparsity ratio $k\%$ or use the same sparsity ratio with reduced bottleneck dimension for the \textit{only mask }setting. We also slightly increase the bottleneck dimension of the \textit{only share} set up to compensate for the parameters of masks. 
Our results, as presented in Table \ref{tab:ablation}, \textcolor{black}{indicate that the random mask setting yields the poorest performance. Moreover,} neither \textit{only mask} nor \textit{only share} reaches the same performance with \propet, highlighting the necessity to use \textbf{both} masking and sharing to achieve higher parameter efficiency.\footnote{Detailed setup can be found in Appendix \ref{app: ablation sec}.}
We believe masking injects crucial structural information into the sub-networks, while sharing is necessary to expand the size of each sub-network when the number of fine-tunable parameters is fixed.  
Therefore, our method can use the parameters more efficiently.

 \label{ablation}
\begin{table}[t!]
\small
\setlength\tabcolsep{3pt}
\centering
\scalebox{0.75}{
\begin{tabular}{l c c  c c c}
\toprule
 Method & \textbf{\propet} & Random & \pbox{3cm}{Only Mask \\ (Same bn)\vspace{0.1em}}  & \pbox{3cm}{Only Mask \\ (Same $k\%$)\vspace{0.1em}} & Only Share \\ 
\midrule
\rowcolor{white!20}\multicolumn{6}{c}{\textbf{GLUE}}\\
\midrule

Adapter (0.11\%) &   \textbf{86.60} &  82.29 &  85.40 &  84.70 &   84.32\\
Prefix (0.11\%)  &    \textbf{84.53} & 79.56 &  84.18 &  84.23&   81.57 \\
LoRA (0.11\%) &   \textbf{85.37} & 82.48  & 83.46 &  84.75 &    82.53    \\
\midrule
\rowcolor{white!20}\multicolumn{6}{c}{\textbf{Ro-En}}\\
\midrule
Adapter (0.46\%) &  \textbf{32.63} & 30.02 & 31.58  & 30.68 & 31.30 \\


\bottomrule
\end{tabular}
}
\caption{Ablation studies of the shared network and masks. We report the average score on GLUE based on \robertabase under single-task learning. For Ro-En, we report the BLEU score with \basebase as the backbone. \textcolor{black}{Numbers in parenthesis indicate the percentage of task-specific bit-level storage calculated against the fully-finetuned model.}}
\label{tab:ablation}

\end{table}

\begin{figure}[t!]
    \centering
    \hspace*{-0.2cm}
    \includegraphics[width=0.48\textwidth]{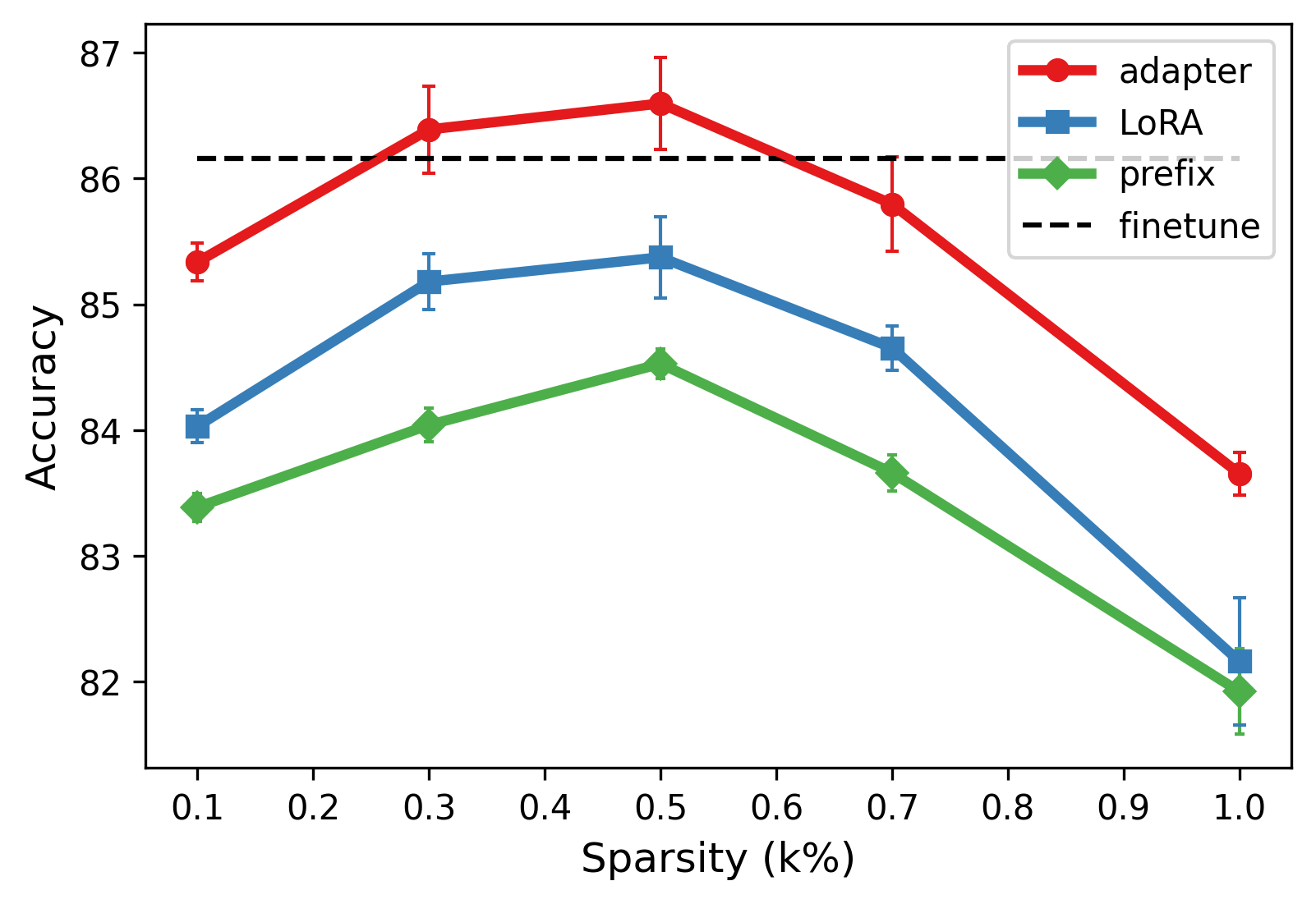}
    \caption{Average score of \propet on GLUE with different sparsity ratios under single-task learning on \robertabase.}
    \label{fig:sparsity}

\end{figure}

\paragraph{How do Sub-Networks' Size and Structure Affect Each Other?}\label{sparsity}
The sparsity ratio $k\%$ is an important hyperparameter in \propet. We study the impact of such sub-network sparsity and present the results in Figure \ref{fig:sparsity}.  
The performance of our \propet improves as $k\%$ increases from 0.1 to 0.5 but then declines as $k\%$ continues to grow from 0.5 to 1.0.
Additionally, it can be seen that all these PETL methods can achieve the best accuracy when $k\%$ is set to 0.5.
This is likely due to the fact that as the network becomes denser from 0.1 to 0.5, the sub-networks get to use more parameters and thus obtain better modeling ability.
However, after 0.5, the network in different layers becomes more homogeneous as the sub-networks overlap more, leading to less distinctive structural information in each layer. The absence of enough structural information starts to harm the performance, which even outweighs the benefits of potential knowledge sharing across the layers. 
We also discover that sharing the PETL module without any pruning ($k\%$ = 1.0) results in the worst performance among all sparsity levels. These results suggest that given the fixed percentage of tunable parameters, it is crucial to find a good balance between the distinctive structural information of each sub-network and the number of parameters used in each sub-network.



\paragraph{How is \propet Conceptually Related to PETL?}
Other than the explanation of prototype network sharing and masking, our proposed \propet can also be considered as a PETL-on-PETL approach, which we refer to as PETL$^2$.
In other words, the mask is to the prototype network (in our approach) as the PETL module is to the PLM (in conventional PETL approaches).
Vanilla PETL methods, such as adapter and prefix-tuning, update specific additional parameters for each  layer and downstream task while only sharing the parameters of the PLM.
In contrast, \propet extends this approach by sharing not only the PLM but also the prototype PETL module among layers and tasks, resulting in a higher degree of parameter sharing.
Our method uses binary masks that function like PETL modules on top of the prototype PETL module to prune different structures in different sub-networks.
These task-specific tunable parameters are thus an order of magnitude smaller than conventional PETL modules.

\section{Conclusion and Future Work}
In this paper, we introduce \propet, a method for sharing prototype PETL modules across different layers and tasks.
Our method significantly improves the parameter efficiency by utilizing the prototype network and maintaining a binary mask for each layer and task.
Extensive experiments show that our method achieves comparable performance to fully fine-tuned models and conventional PETL methods with a much smaller fraction of storage.
For future works, we aim to study the interpretability of the masks in different layers and explore their potential relationships.
We also intend to apply our method to the pre-training process of large language models to reduce the overall number of parameters.

\section*{Limitations}
Although our masks in different layers are binary and require significantly less storage compared to other PETL networks, we still need the underlying 32-bit scores for each mask during the training process. Therefore, \propet consumes slightly more memory in the training time than existing PETL methods. 
To fine-tune \propet, it takes a similar training time to conventional PETL modules, which means our method will normally take a longer time to converge compared to the fully fine-tuned model.




\section*{Acknowledgements}
We would like to thank the anonymous reviewers, our meta-reviewer, and senior area chairs for their constructive comments and support on this work.
This research/project is supported by the National Research Foundation Singapore and DSO National Laboratories under the AI Singapore Program (AISG Award No: AISG2-RP-2020-016), and AI Singapore Programme (AISG Award No: AISG2-PhD-2021-08-007).


\bibliography{anthology}
\bibliographystyle{acl_natbib}

\appendix
\section{Implementation Details of the Three Variants of \propet} \label{app:PETL module}
\subsection{\propetadapter} An adapter module modifies a model's hidden representation through a down-sampling projection and an up-sampling projection with a non-linear layer in between:
 \begin{equation}
        \boldsymbol{h} \leftarrow \boldsymbol{h} + f( \boldsymbol{h} W_{\text{down}} + b_{\text{down}})W_{\text{up}} + b_{\text{up}}
\end{equation}
where $W$ represents the weight matrix, $f$ denotes the non-linear layer and $b$ is the bias term. In \propetadapter, we apply our binary pruning masks on the up-sampling and down-sampling weights, respectively:
 \begin{equation}
        \boldsymbol{h} \leftarrow \boldsymbol{h} + f( \boldsymbol{h} \widetilde{W}_{\text{down}} + b_{\text{down}})\widetilde{W}_{\text{up}} + b_{\text{up}}
\end{equation}
where $\widetilde{W}_{\text{down}} = W_{\text{down}} \odot m_{\text{down}}$ and $\widetilde{W}_{\text{up}} = W_{\text{up}} \odot m_{\text{up}}$.

The original Houlsby adapter~\cite{HoulsbyGJMLGAG19} introduces the adapter module after each multi-head attention and feed-forward layer. \citet{PfeifferKRCG21} later propose a more efficient variant of the adapter that is only inserted after the feed-forward layer. 

\
\subsection{\propetlora}
LoRA~\cite{HuSWALWWC22}, like adapter tuning, also includes a down-sampling projection and an up-sampling projection. The difference is that LoRA does not have any non-linear layers, but it does have an additional scaling factor $\alpha \geq 1$:
\begin{equation}
        \boldsymbol{h} \leftarrow \boldsymbol{h} + \alpha \cdot \boldsymbol{x} W_{\text{down}}W_{\text{up}}
\end{equation}
To modify PLMs' hidden representation, LoRA is applied to the query and value representations of the attention modules in Transformers. In \propetlora, the network is pruned with binary masks:
\begin{equation}
\begin{aligned}
        \boldsymbol{h} \leftarrow \boldsymbol{h} + \alpha \cdot \boldsymbol{x} (W_{\text{down}}\odot m_{\text{down}})(W_{\text{up}}\odot m_{\text{up}})
\end{aligned}
\end{equation}

The scaling factor $\alpha$ is an important parameter for LoRA and \citet{HuSWALWWC22} use different $\alpha$ for different datasets in their released code. 
In \propetlora, applying the pruning masks $m$ will reduce the norm of the output of the LoRA module. We find that applying a larger $\alpha$ for \propetlora than that used in LoRA will remedy the issue of reduced feature norm and result in better performance.

\subsection{\propetprefix}\label{app:prefix-tuning}
\citet{LiL20} propose to insert tunable matrices, which they call prefixes, into the key-value pair of the Transformers' attention modules:
\begin{equation}
\begin{aligned}
H &= \attention(Q, [P_k;K], [P_v;V]) \\&= \softmax(\frac{Q[{P}_k;K]^T}{\sqrt{d}})[P_v;V]
\end{aligned}
\end{equation}
where $[\cdot ; \cdot]$ stands for matric concatenation. They also find that directly updating the $P$ parameters will lead to unstable training and a bit drop in performance.
In order to ameliorate the problem, they propose to reparametrize the matrix:
\begin{equation}
\begin{aligned}
P&=P'W
\end{aligned}
\end{equation}
where  $P'$ is a smaller learnable matrix and $W$ is the weight of an up-projecting feedforward neural network.
In \propetprefix, we apply our binary pruning masks on the reparametrized prefixes:
\begin{equation}
\begin{aligned}
H &= \attention(Q, [P_k \odot m_k;K], [P_v \odot m_v;V])
\end{aligned}
\end{equation}
Note that different from adapter and LoRA tuning, the pruning masks in \propetprefix do not directly operate on the parameters of the network. Instead, the masks are applied on $P_k$ and $P_v$ that are output activation depending collectively on $W$ and $P'$. Thus, it might be hard for the mask training process to identify good structures from  $P_k$ and $P_v$, which potentially explains the sub-optimal results of \propetprefix in Table \ref{tab:single}. To verify this claim, we further compare 3 PETL modules against their counterparts pruned with a sparsity of 0.5 in Table \ref{tab: explain prefix}. We find that both adapter and LoRA tuning improve their performance when 50\% of parameters are pruned, which substantiates the idea that structural information of sub-networks is important. However, for prefix tuning that uses reparameterization, its performance drops when we learn pruning masks on top of it. This shows that the masks fail to locate suitable structures for prefix tuning.

\begin{table}[t!]
\small
\setlength\tabcolsep{6.1pt}
\centering
\scalebox{0.93}{
\begin{tabular}{l  c  c}
\toprule
Module & Vanilla & Prune $50\%$ \\ 
\midrule

Prefix (l=64) &    \textbf{84.97} & 84.50  \\ 
LoRA (bn=32)&  84.96 & \textbf{85.44} \\ 
Adapter (bn=64)&   85.65 & \textbf{86.56}  \\ 



\bottomrule
\end{tabular}
}
\caption{Average score between pruning or not on the GLUE dataset based on \robertabase single-task learning under conventional PETL methods.}
\label{tab: explain prefix}
\end{table}

\section{Parameter Efficiency in \propet}\label{app:efficiency}
\subsection{Overview}
    \begin{table}[ht]
    \large
\setlength\tabcolsep{4.5pt}
    \centering
    \scalebox{0.7}{
    \begin{tabular}{ll} \toprule
    Learning Setup  & \pbox{3cm}{Bit-Level Storage\\ of \propet} \\ 
    \midrule
    Single-task Learning & $p + 1/32  p  L$  \\
    Multi-task Learning & $p + 1/32 p  (L + T)$ \\ 
    \bottomrule
    \end{tabular}}
    \caption{Storage calculation.}
    \label{table:parameter_calculation}
    \end{table}

We present an approximate calculation of the bits (space) required by \propet during inference in Table~\ref{table:parameter_calculation}, in which we assume that a single shared PETL network contains $p$ BLS and the model has $L$ layers.
In addition, the storage space required for the binary mask makes up $1/32$ ($\approx 0.031$) of the 32-bit PETL module. Depending on the specific PETL module used, the calculations may vary a bit, and we detail them in the following sections.

\subsection{\propetadapter}
Given the number of layers ($L$), hidden dimension ($d$) of the pre-trained language model, and bottleneck dimension ($bn$) of the adapter module, \textcolor{black}{the  BLS consumed} by \propetadapter is:
\begin{equation}
\begin{aligned}
\underbrace{32 \cdot (2\cdot bn\cdot d + bn + d)}_{\text{Prototype} \; \text{adapter}} + \underbrace{2 \cdot bn \cdot d \cdot L}_{\text{Mask}}
\end{aligned}
\end{equation}
Note that our method does not apply any masks on the bias terms in the prototype adapter. 
Since our \propet will reuse the prototype network, we do not consider the pruning ratio $k\%$ when calculating the \textcolor{black}{storage}.
Here we also discuss how to calculate the \textcolor{black}{storage cost} of parameters used under the \textit{only mask} setting in Table \ref{tab:ablation}, in which we do not share the PETL module across layers but only learn masks to prune them.
In that case, we do not count the pruned parameters because they will never be reused.
Therefore, the formula for the  BLS required by \textit{only mask} is:
\begin{equation}
\begin{aligned}
32 \cdot (2\cdot k\% \cdot bn\cdot d  + bn + d) \cdot L
\end{aligned}
\end{equation}

\begin{table*}[ht!]
\normalsize
\setlength\tabcolsep{6pt}
\centering
\scalebox{0.8}{
\begin{tabular}{l cccccccccc}
\toprule
{Datasets}& XSum & Ro-En& CoLA & SST-2 & MRPC & QQP & STS-B & MNLI & QNLI & RTE\\
\midrule
\#Train& 204k& 997k	 & 8.6k	 & 66k & 3.7k & 363k  & 5.7k & 392k & 104k & 2.5k \\
\#Valid& 11k& 2.6k	 & 0.5k	 & 1k  & 0.2k   & 1k  & 0.8k  & 1k & 1k & 0.1k\\
\#Test&  11k  & 3k    & 0.5k    &0.9k & 0.2k & 40k & 0.8k   & 10k & 5k & 0.1k\\
\bottomrule
\end{tabular}
}
\caption{Dataset statistics}
\label{tab:statistics}
\end{table*}

\subsection{\propetlora}
Similarly, given the number of layers ($L$), hidden dimension ($d$) of the pre-trained language model, and bottleneck dimension ($bn$) of the LoRA module,  the formula for the  storage needed by \propetlora is:
\begin{equation}
\begin{aligned}
\underbrace{32 \cdot 4\cdot bn\cdot d}_{\text{Prototype} \; \text{LoRA}} + \underbrace{4 \cdot bn \cdot d \cdot L}_{\text{Mask}}
\end{aligned}
\end{equation}
Note that there is no bias term in LoRA. Under the \textit{only mask} setup, given the sparsity ratio $k\%$, the  \textcolor{black}{storage} of parameters is:
\begin{equation}
\begin{aligned}
32 \cdot 4\cdot k\% \cdot bn\cdot d \cdot L
\end{aligned}
\end{equation}

\subsection{\propetprefix}
Given the number of layers ($L$), hidden dimension ($d$) of the pre-trained language model, and the prefix length ($l$) of the prefix module, the formula to calculate the BLS of \propetprefix is:
\begin{equation}
\begin{aligned}
\underbrace{32 \cdot 2\cdot l\cdot d}_{\text{Prototype}\; \text{prefix}} + \underbrace{ 2 \cdot l \cdot d \cdot L}_{\text{Mask}}
\end{aligned}
\end{equation}
Under \textit{only mask} settings, given the sparsity ratio $k\%$, the  \textcolor{black}{approximate required  storage} is:
\begin{equation}
\begin{aligned}
2\cdot k\% \cdot l\cdot d \cdot L
\end{aligned}
\end{equation}

\section{Experimental Details}\label{app:experiment details}
We briefly introduce the benchmark datasets used in this work. Their statistics can be found in Table~\ref{tab:statistics}.
\subsection{Datasets}
\paragraph{GLUE} The General Language Understanding Evaluation (GLUE) benchmark~\cite{wang-etal-2018-glue} is widely used to benchmark models' language understanding ability. It consists of a broad range of sentence-level tasks, including natural language inference (MNLI, QNLI, RTE), sentence similarity (MRPC), paraphrase detection (QQP, STS-B), and single sentence classification (SST-2, CoLA).
\paragraph{XSum} The Extreme Summarization dataset (XSum)~\cite{NarayanCL18}  is designed to evaluate systems' abstractive summarization ability of a single document. It is collected from the online articles of the British Broadcasting Corporation (BBC). The input is a single document with an average token count of 431.07, and the model is expected to generate a short, single-sentence summarization of this document.
\paragraph{WMT16 Ro-En}  WMT16 is the 2016 edition of the Workshop on Machine Translation (WMT) dataset~\cite{wmt16}. This dataset is widely used to evaluate machine translation systems. To benchmark on WMT16 Ro-En, the model is supposed to take in a Romanian sentence and output translation in English.

\begin{table}[ht]
\small
\setlength\tabcolsep{3pt}
\centering
\scalebox{0.95}{
\begin{tabular}{l   c }
\toprule
 Model &     Mask\textsubscript{lr}  \\
\midrule
\propetadapter &  3e-3\\
\propetlora & 3e-2\\
\propetprefix  & 1e-4\\
\bottomrule
\end{tabular}
}
\caption{Learning rates in GLUE under single task settings}
\label{tab:lr}
\end{table}

\begin{table}[ht]
\small
\setlength\tabcolsep{3pt}
\centering
\scalebox{0.95}{
\begin{tabular}{l c c cccccc}
\toprule
 Datasets &  Training time\\
\midrule
 MNLI & 90min\\
 QQP & 81min \\
 QNLI & 24min\\
 SST-2 & 15min\\
 CoLA & \textcolor{white}{0}4min\\
 STS-B & \textcolor{white}{0}3min \\
 MRPC & \textcolor{white}{0}2min \\
 RTE & \textcolor{white}{0}1min\\

\bottomrule
\end{tabular}
}
\caption{The approximate training time in GLUE 
with \propet under single-task training}
\label{tab:tt}
\end{table}

\subsection{Implementation Details}

\paragraph{Single-Task Learning on \robertabase}
Our  models are implemented based on the AdapterHub package~\cite{adapterhub}.
We use the \textit{datasets}~\cite{lhoest-etal-2021-datasets} library to calculate  each sub-task's scores in GLUE.

The learning rate of the PETL module is set to 1e-4, and we detail the learning rate of pruning masks  in Table~\ref{tab:lr}. 
We find that it is important to set a higher learning rate for the masks than the learning rate of the PELT network.
The batch size is set to 128 and the weight decay is 0.1 in all experiments with Roberta.
When conducting experiments on GLUE datasets, we train 10 epochs when the dataset is large (MNLI, QNLI, QQP, SST-2), while training 20 epochs on the small dataset (RTE, MRPC, STS-B, CoLA).
Table~\ref{tab:tt} shows the training time taken on a single A100 GPU.

\paragraph{Single and Multi-Task Learning on \basebase}
We implement T5 based on the transformers library~\cite{wolf-etal-2020-transformers}. We use the ROUGE package~\cite{lin-2004-rouge} for ROUGE-2 calculation and sacrebleu~\cite{post-2018-call} for the BLEU score. Table~\ref{tab:xsum hyper-param} shows the training time on a single A100 GPU and the detailed hyperparameters used. We mainly follow \citet{HeZMBN22} and \citet{MahabadiR0H20} to select the hyperparameters and do not perform an exhaustive search. The same set of hyperparameters is used across the fully-finetuning, adapter tuning, and \propet models.   For GLUE, we follow \citet{MahabadiR0H20} to sample data from each GLUE sub-task with a temperature of 10. Specifically, a sub-task is sampled with probability  $p_{\tau}^{1/T}$ where $p_{\tau} = \frac{N_{\tau}}{\sum_{i=1}^T N{\tau}}$ and $ T = 10$.
\begin{table}[ht!]
\small
\setlength\tabcolsep{3pt}
\centering
\scalebox{0.95}{
\begin{tabular}{l   c c c}
\toprule
 Hyperparameters &     XSum  & Ro-En & GLUE\\
\midrule
 Batch size&  64&  64 & 128\\
 Total steps & 100k & 60k & 20k\\
Learning rate & 1e-4 & 3e-4 & 3e-4\\
Mask's learning rate & 1e-3 & 3e-3 & 3e-3\\
 Learning rate schedule & linear& linear & linear\\
Label smoothing & 0.1 & 0.1 & 0.1\\
Weight decay & 0.01& 0.01 & 0.0\\
Sampling temperature & n.a. & n.a. & 10 \\
\midrule
Training time & 1 day & 6 hours & 2 hours \\
\bottomrule
\end{tabular}
}
\caption{Hyperparameters on XSum, Ro-En, and GLUE when \basebase is used as the backbone.}
\label{tab:xsum hyper-param}
\end{table}

\begin{table}[ht!]
\small
\setlength\tabcolsep{3pt}
\centering
\scalebox{0.95}{
\begin{tabular}{l   c }
\toprule
Combining Method &     Avg. GLUE Score \\
\midrule
OR &  \textbf{85.97}\\
ADD & 85.66\\
AND  & 85.57\\
\bottomrule
\end{tabular}
}
\caption{Results of different mask combining methods based on T5 multi-task learning on GLUE. 
We set the bottleneck dimension to 64.}
\label{tab:maskablate}
\end{table}

\begin{table}[ht!]
\small
\setlength\tabcolsep{4pt}
\centering
\scalebox{0.7}{
\begin{tabular}{l cc  cc  c c}
\toprule
 Method  & \propet & \pbox{3cm}{Only Mask \\ (Same bn)\vspace{0.1em}}  & \pbox{3cm}{Only Mask \\ (Same $k\%$)\vspace{0.1em}} & \pbox{3cm}{ \, \, \, Only Mask \\ (Same $k\%$ and bn)\vspace{0.1em}}  & Only Share \\ 
\midrule
\rowcolor{white!20}\multicolumn{6}{c}{\textbf{GLUE}}\\
\midrule

Adapter & 64/50\%  & 64/11\%& 14/50\% &64/50\% & 88/100\%\\
Prefix &  64/50\% &     64/15\%& 15/50\% &64/50\% & 88/100\%\\
LoRA &  32/50\% &   32/15\%& 8/50\% & 32/50\%& 44/100\%\\
\% \textcolor{black}{BLS} &  (0.11\%)  &  (0.11\%) &  (0.11\%) & (0.48\%) &  (0.11\%)\\
\midrule
\rowcolor{white!20}\multicolumn{6}{c}{\textbf{Ro-En}}\\
\midrule

Adapter & 384/50\% &  384/8\% &55/50\% & 384/50\%& 673/100\% \\
\% \textcolor{black}{BLS} &  (0.46\%)  & (0.46\%)  &  (0.46\%) & (3.2\%)& (0.46\%)\\
\bottomrule
\end{tabular}
}
\caption{The bottleneck dimension/sparsity ratio used in Table \ref{tab:app:ablation}.}
\label{tab:app:hyper}

\end{table}
\begin{table}[ht!]
\small
\setlength\tabcolsep{4pt}
\centering
\scalebox{0.7}{
\begin{tabular}{l c  c c c c }
\toprule
 Method & \textbf{\propet} &  \pbox{3cm}{Only Mask \\ (Same bn)\vspace{0.1em}}  & \pbox{3cm}{Only Mask \\ (Same $k\%$)\vspace{0.1em}} &\pbox{3cm}{ \, \, \, Only Mask \\ (Same $k\%$ and bn)\vspace{0.1em}}  & Only Share \\ 
\midrule
\rowcolor{white!20}\multicolumn{6}{c}{\textbf{GLUE}}\\
\midrule

Adapter &   \textbf{86.60} &  85.40 &  84.70 & 86.56&   84.32\\
Prefix  &    \textbf{84.53} & 84.18 &  84.23&  84.50 & 81.57 \\
LoRA &   85.37 & 83.46 &  84.75 &   \textbf{85.44} & 82.53    \\
\% \textcolor{black}{BLS} &  (0.11\%)  &  (0.11\%) &  (0.11\%) & (0.48\%) &  (0.11\%)\\
\midrule
\rowcolor{white!20}\multicolumn{6}{c}{\textbf{Ro-En}}\\
\midrule

Adapter &  32.63 & 31.58  & 30.68 & \textbf{33.28} & 31.30 \\
\% \textcolor{black}{BLS} &  (0.46\%)  & (0.46\%)  &  (0.46\%) & (3.2\%)& (0.46\%)\\
\bottomrule
\end{tabular}
}
\caption{Ablation studies of the shared network and masks. We report the average score on GLUE based on \robertabase under single-task learning. For Ro-En, we report the BLEU score with \basebase as the backbone.}
\label{tab:app:ablation}

\end{table}

\section{Additional Ablation Studies}
\subsection{Choice of Mask Combining Methods} \label{mask_combine_ablat}
As shown in Table \ref{tab:maskablate}, using the OR logical operation to combine the layer mask and the task mask achieves the best performance. This is intuitive because given a specific task and a Transformer layer, parameters that contain the layer information and parameters that contain the task information should be \textbf{both} used in the forward pass.

\begin{table*}[h!]
\small
\setlength\tabcolsep{3.4pt}
\centering
{
\scalebox{0.9}{
\begin{tabular}{l |c c c  c c c c  }
\toprule

\rowcolor{white!20}\multicolumn{8}{l}{\it \textbf{Adapter}}\\

\midrule
Bottleneck dimension &12 & 24 & 48 & 96 & 192 & 384 & 769   \\

\% Bit-Level Storage&    0.207\% & 0.405\% & 0.803\% & 1.597\% & 3.186\% & 6.363\% & 12.72\% \\
ROUGE-2 
& 14.33 & 15.05 & 15.98 & 16.66 & 16.97 & 18.29 & 18.58 \\
\midrule
\rowcolor{white!20}\multicolumn{8}{l}{\it \textbf{\propetadapter}}\\
\midrule
Bottleneck dimension & 96 & 192 & 384 & 768 & 1536 & 3042 & 6144\\
\% Bit-Level Storage&   0.116\% & 0.232\%  & 0.464\%  &  0.927\%  & 1.853\%  & 3.706\%  & 7.412\%  \\
ROUGE-2 &  15.52 & 16.42 & 16.96 & 17.91 &   18.52 &  18.87 &  18.96 \\ 
\bottomrule
\end{tabular}
}
}
\caption{Performance of adapter and \propetadapter on XSum under single-task learning when varying the bottleneck dimension. We report the BLEU and the proportion of the bit-level storage.}
\label{tab: Xsum}
\end{table*}

\begin{table*}[h]
\small
\setlength\tabcolsep{3.4pt}
\centering
{
\scalebox{0.9}{
\begin{tabular}{l |c c c  c c cc  c }
\toprule

\rowcolor{white!20}\multicolumn{9}{l}{\it \textbf{Adapter}}\\

\midrule
Bottleneck dimension & 6 & 12 & 24 & 48 & 96 & 192 & 384 & 768  \\

\% Bit-Level Storage&   0.108\% & 0.207\% & 0.405\% & 0.803\% & 1.597\% & 3.186\% & 6.363\% & 12.72\%  \\

BLEU&26.95 & 29.18 & 30.20 & 31.20 & 32.52 & 32.83 & 33.56 & 33.63  \\
\midrule
\rowcolor{white!20}\multicolumn{9}{l}{\it \textbf{\propetadapter}}\\
\midrule
Bottleneck dimension & -- & 96 & 192 & 384 & 768 & 1536 & 3072 & 6144 \\
\% \textcolor{black}{Bit-Level Storage} & --  &0.116\% & 0.232\% & 0.464\% & 0.927\% & 1.853\% & 3.706\% & 7.412\% \\
BLEU  & --  & 30.82 & 32.72 & 32.63 & 33.16 & 33.62 & 33.79 & 33.83 \\
\bottomrule

\end{tabular}
}
}

\caption{Performance of adapter and \propetadapter on En-Ro under single-task learning when varying the bottleneck dimension. We report the BLEU and the proportion of the bit-level storage.}
\label{tab: EnRo}
\end{table*}

\begin{table*}[ht!]
\small
\setlength\tabcolsep{3.4pt}
\centering
{
\scalebox{0.73}{
\begin{tabular}{l |c c c c c c c c c c c c c c c  }
\toprule

\rowcolor{white!20}\multicolumn{16}{l}{\it \textbf{Adapter}}\\
\midrule
Bottleneck dimension & --& -- & -- & -- & 1 & 2 & 3 & 6 & 12 & 24 & 48 & 64 & 96 & 192 & 384    \\

\% BLS per task & --& -- & -- & -- &   0.003\%
& 0.005\%
& 0.007\%
& 0.013\%
& 0.026\%
& 0.051\%
& 0.100\%
& 0.133\%
& 0.200\%
& 0.398\%
& 0.795\% \\
Avg. Score & --& -- & -- & -- & 80.88 & 82.61 & 82.89 & 83.94 & 84.92 & 85.00 & 85.49 & 85.48 & 85.78 & 86.01 & 85.50 \\

\midrule
 \rowcolor{white!20}\multicolumn{16}{l}{\it \textbf{\propetadapter}}\\
\midrule
Bottleneck dimension  & 1 & 2 & 3 & 6 & 12 & 24 & 48 & 64 & 96 & 192 & 384 & 768 & 1536 & 3072 & --    \\
\% \textcolor{black}{BLS per task}

& 0.0002\%
& 0.0004\%
& 0.0006\%
& 0.001\%
& 0.002\%
& 0.004\%
& 0.008\%
& 0.011\%
& 0.017\%
& 0.033\%
& 0.066\%
& 0.132\%
&  0.265\%
&  0.529\%
& -- \\ 
Avg. Score 

& 79.88
& 83.45
&  84.46
& 85.32
& 85.77
& 85.85
& 85.93
& 85.97
& 85.76
& 86.05
& 85.73
& 86.3
& 86.21
& 85.71
& --  \\ 
\bottomrule
\end{tabular}
}}

\caption{Performance of adapter and \propetadapter on GLUE under multi-task learning when varying the bottleneck dimension. We report the average score and the proportion of the bit-level storage per task.}

\label{tab: GLUET5}
\end{table*}

\subsection{Choice of Sharing and Masking} \label{app: ablation sec}
We provide additional details of the experiments in Table \ref{tab:ablation} in this section.
 With the equations shown in Appendix \ref{app:efficiency}, we calculate the bottleneck dimension ($bn$) and sparsity ratio ($k\%$) to ensure all the settings have a similar \textcolor{black}{BLS} to \propet. 
We also add one more setup, \textit{only mask} with the same $k\%$ and $bn$ as \propet. 
Note that this new setup will result in more \textcolor{black}{storage} than the other settings. The hyperparameters are listed in Table~\ref{tab:app:hyper} and the results are detailed in Table \ref{tab:app:ablation}.

We find that on the GLUE dataset, \propet achieves matched performance to this new setup, even though the latter has used more than 4 times the \textcolor{black}{bit-level storage} (0.11\% v.s. 0.49\%).
We believe that, with masks, \textcolor{black}{models with only 0.11\% of  storage} can be enough to have a good performance while more \textcolor{black}{storage cost} will not have a significant improvement or lead to overfitting under simple tasks like GLUE.
However, on the more challenging dataset Ro-En, when we keep the $k\%$ and $bn$ unchanged and only mask the adapter modules, the model indeed improves its performance  by 0.65, but this comes at a cost of around 7$\times$ more \textcolor{black}{storage of parameters}.
\section{Additional Results}\label{hugetable}
We additionally present the experimental results from Figure~\ref{dist} in table format in Table~\ref{tab: Xsum},~\ref{tab: EnRo}, and~\ref{tab: GLUET5}.

\end{document}